\documentclass{article}

\usepackage{graphicx}
\usepackage{multirow}
\usepackage{comment}

\PassOptionsToPackage{numbers, compress}{natbib}

\usepackage[final]{DLI_2023}

\usepackage[utf8]{inputenc} 
\usepackage[T1]{fontenc}    
\usepackage{hyperref}       
\usepackage{url}            
\usepackage{booktabs}       
\usepackage{amsfonts}       
\usepackage{nicefrac}       
\usepackage{microtype}      
\usepackage{xcolor}         

\title{Harnessing Transfer Learning from Swahili: Advancing Solutions for Comorian Dialects}

\usepackage{authblk}

\author[1,2]{Abdou Mohamed Naira}
\author[1]{Zakarya Erraji}
\author[1,2]{Abdessalam Bahafid}
\author[1,2]{Imade Benelallam}

\affil[1]{Institut National de Statistique et d'Economie Appliquée, Rabat, Morocco}
\affil[2]{ToumAI Analytics, Rabat, Morocco}

\begin{document}

\maketitle

\begin{abstract}
	If today some African languages like Swahili have enough resources to develop high-performing Natural Language Processing (NLP) systems, many other languages spoken on the continent are still lacking such support. For these languages, still in their infancy, several possibilities exist to address this critical lack of data. Among them is Transfer Learning, which allows low-resource languages to benefit from the good representation of other languages that are similar to them. In this work, we adopt a similar approach, aiming to pioneer NLP technologies for Comorian, a group of four languages or dialects belonging to the Bantu family.

	Our approach is initially motivated by the hypothesis that if a human can understand a different language from their native language with little or no effort, it would be entirely possible to model this process on a machine. To achieve this, we consider ways to construct Comorian datasets mixed with Swahili. One thing to note here is that in terms of Swahili data, we only focus on elements that are closest to Comorian by calculating lexical distances between candidate and source data. We empirically test this hypothesis in two use cases: Automatic Speech Recognition (ASR) and Machine Translation (MT). Our MT model achieved ROUGE-1, ROUGE-2, and ROUGE-L scores of 0.6826, 0.42, and 0.6532, respectively, while our ASR system recorded a WER of 39.50\% and a CER of 13.76\%. This research is crucial for advancing NLP in underrepresented languages, with potential to preserve and promote Comorian linguistic heritage in the digital age.

\end{abstract}

\section{Introduction}

Natural Language Processing (NLP) solutions have a wide range of applications that are increasingly integrated into the tools we use to facilitate our daily lives. These applications span from simple Machine Translation (MT) to more complex tasks such as medical diagnostics and Spam Detection \cite{Khurana2022}. Given their advantages, it is essential to democratize these solutions across different languages and countries so that everyone can benefit from them. For instance, NLP solutions like MT could enhance the tourist experience by facilitating communication with locals in a country like the Comoros archipelago, where the number of tourists has significantly increased in recent years \cite{ComoresA0:online}.

The Comorian language consists of four variants, each one is primarily spoken on one of the islands of the archipelago. There is ShiNgazidja spoken on the island of Ngazidja, ShiMwali spoken in Moheli, ShiNdzuani spoken in Anjouan, and ShiMaore in Mayotte. However, as Ahmed-Chamanga continually demonstrates in \cite{chamanga}, this diversity of Comorian may not fully reflect reality, given that there are also variations within each island. For example, a native of Mbeni in the north of Ngazidja might have difficulty understanding certain words when talking with a native of Foumbouni in the south of the same island. One can cite the example of the word "egg" which translates in ShiNgazidja to "djwai" ("madjwai" in plural) but also translates to "dzundzu" a common term in some regions but entirely unknown in others. Some even use the term "mayayi" for the plural. Ideally, it would be interesting to consider this diversity when constructing NLP systems involving Comorian. However, the challenge of obtaining granular data specific to each variant necessitates resorting to mixed solutions where Comorian is treated as a whole.

In this study, we undertake an exercise to explore how, starting with a small dataset, we can develop NLP solutions for an extremely low-resource language. We employ a transfer learning approach between two closely related languages: a well-resourced source language and an extremely low-resource target language. The goal is to investigate how the former can assist in creating one of the first datasets for the latter. We apply this approach with Comorian as the target language and Swahili as the source language. More specifically, the main contributions of this work are:

\begin{itemize}
	\item \textbf{Datasets}: We create datasets by combining existing Swahili corpora, filtered for their lexical proximity to Comorian, with local Comorian data. This process includes cleaning and normalizing textual data, as well as processing audio data, to build robust datasets suitable for training NLP models for Comorian. 
	
	\item \textbf{Models}: We develop and evaluate an ASR model that was initially pre-trained on Swahili data, and subsequently fine-tune it for Comorian. Following a similar approach, we fine-tune a pre-trained MT model. These models demonstrate that effective NLP solutions can be created for Comorian and other low-resource languages, even when starting with extremely limited resources.

\end{itemize}

\section{Related Work} \label{related_work}
Very few works have been carried out previously in the field of NLP on Comorian. However, there are two notable works focusing on Comorian that we can mention here. Both aimed to initiate the representation of this language in the NLP domain by proposing dataset construction solutions:

\begin{itemize}
	\item \textbf{MT Dataset} \cite{abdourahamane:hal-01992871}: Here, the goal was to build a dataset for machine translation from Comorian to French using Swahili as a pivot language. The solution involved initially retrieving articles written in French and then translating them into Swahili using an open-source machine translation system. From there, manual rewriting was done from Swahili to Comorian. Using Swahili as a pivot helped reduce labeling effort due to the similarity between the two languages, instead of directly labeling from French.
	
	\item \textbf{Datasets for multiple domains} \cite{naira-etal-2024-datasets}: This work aimed to propose several datasets by monitoring all existing resources (lexicon, raw text, audio, etc.). The constructed datasets were evaluated in five tasks: Language Identification, Sentiment Analysis, Part-Of-Speech Tagging, and ASR.
\end{itemize}

\section{Lexical Distances} \label{lexical_distance}
The idea here is to determine whether it is truly legitimate to use Swahili as a candidate language for conducting transfer learning on ShiKomori. To do this, we first revisit previous work comparing the similarities between Bantu languages and then, we attempt to quantify the lexical proximity between Comorian dialects and Swahili.

\subsection{Guthrie's Classification}

\begin{figure}
	\centering
	\includegraphics[width=0.5\linewidth]{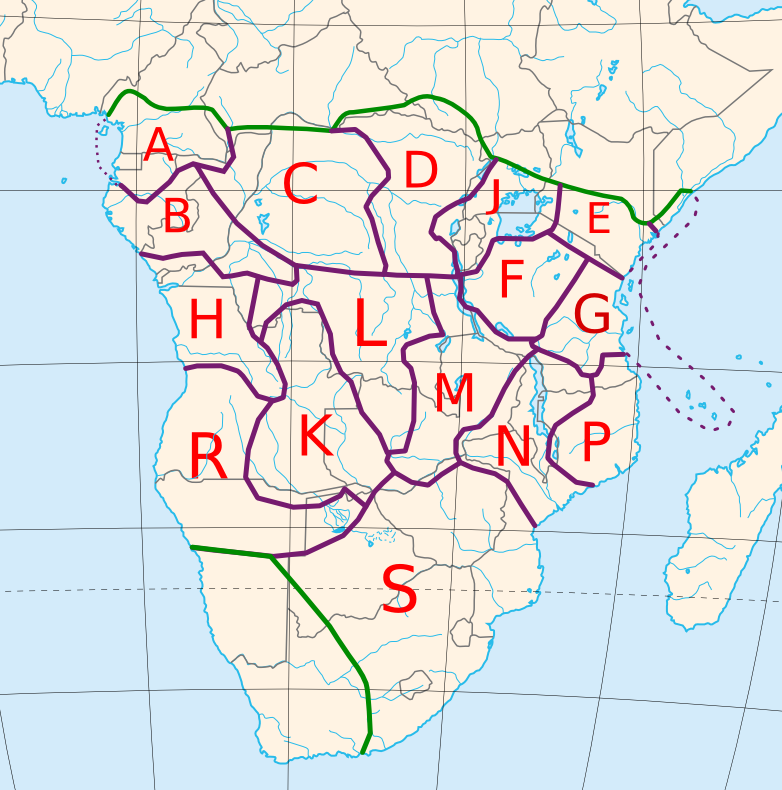}
	\caption{Approximate locations of the sixteen Guthrie Bantu zones.}
	\label{fig:bantu_classification}
\end{figure}

Guthrie's classification of Bantu languages \cite{Smith_1949} is a crucial framework for understanding the diversity and distribution of languages within the Bantu family. This classification divides the Bantu languages into sixteen distinct zones, each grouping languages that exhibit significant similarities in terms of phonology, lexicon, and sometimes grammar. Referring to Figure \ref{fig:bantu_classification}, it is evident that each zone corresponds to a geographic area where languages share common features, reflecting historical influences, migrations, and cultural interactions among linguistic communities.

For instance, in Guthrie's group G, languages such as Swahili and Comorian are included. They display numerous lexical similarities, which is not surprising given their geographical proximity and the historical trade exchanges in the Indian Ocean region. These exchanges fostered linguistic and cultural interdependence, contributing to the formation of a shared vocabulary or reciprocal lexical borrowings.

\subsection{Distance Calculations}
We estimate the lexical distances between these languages using an approach that employs the Levenshtein distance and three variants of the Swadesh list \cite{a3f69dd7-61c8-3f9c-ba53-bc66666038a5, Petroni2010, TheSabak97:online, Beaufils2020}. Equation \ref{eq1} summarizes the distance calculations, where $\alpha_i$ and $\beta_i$ represent two equivalent words in two languages, $D_l(\alpha_i, \beta_i)$ is the Levenshtein distance between them and $L(\alpha_i, \beta_i)$ is the length of the longest word. The distance ranges from 0 to 100, where 0 represents two identical words.

\begin{equation}
	\label{eq1}
	D(\alpha_i, \beta_i) = \frac{D_l(\alpha_i, \beta_i)}{L(\alpha_i, \beta_i)} \cdot 100
\end{equation}

The Swadesh list consists of fundamental words introduced by the linguist Morris Swadesh during the 1950s. This compilation aims to highlight words that are common across various languages. It serves as a tool in comparative linguistics for examining the connections between different languages and for reconstructing ancestral languages. Usually comprising around 100 to 200 essential terms, the list includes items like body part names, pronouns, verbs denoting actions, descriptive words and vocabulary related to family, animals, numbers and the environment. Linguists utilize this list to make comparisons between languages, exploring both their commonalities and unique characteristics.

We construct an expanded Swadesh list comprising 207 words for each ShiKomori dialect as well as for Swahili. This compilation is achieved by sourcing data from the ORELC lexicon and the dictionaries referenced in \cite{naira-etal-2024-datasets}. We use French as our pivot language to align Swahili terms with their corresponding ShiKomori equivalents. For terms not found in the dictionaries, we consult Glosbe\footnote{\url{https://fr.glosbe.com/}}, an online dictionary that enriches the ShiKomori. Despite these efforts, some words remain absent. To address this, we assign a lexical distance by averaging the distances between all existing word pairs in the respective language pair. For instance, if a word is missing in ShiNgazidja, we calculate its distance to its synonym in ShiNdzuani as the average distance between all existing word pairs.

In Table \ref{tab:distance}, we observe that naturally, the four dialects exhibit many similarities. Pairwise similarities align with the conclusion drawn in \cite{TheSabak97:online}, suggesting that the dialects are divided into two groups: ShiNgazidja-ShiMwali and ShiNdzuani-ShiMaore. However, what interests us most here are the distances between the four dialects and Swahili. In fact, we observe slightly higher distances compared to those between the dialects, but they still demonstrate the closeness between them and Swahili. This leads us to conduct experiments considering the following hypotheses in the subsequent sections of this work:
\begin{itemize}
	\item \textbf{Intra-Lingual Transfer}: Due to the impressive similarities among the four dialects, a system capable of supporting them simultaneously would be minimally affected by the few differences or the unique properties that a dialect may have.
	\item \textbf{Cross-Lingual Transfer}: Given the severe lack of data, it would be entirely feasible to use Swahili as a starting point for developing initial systems that include these dialects.
\end{itemize}

\begin{table*}[h]
	\footnotesize
	\caption{Pairwise lexical distances across the three variations of the Swadesh lists.}
	\begin{center}
		\begin{tabular}{@{}lllllll@{}}
			\toprule
			
			Versions & Languages & Swahili & ShiNgazidja & ShiMaore & ShiNdzuani & ShiMwali \\
			\midrule
			\multirow{5}{*}{Swadesh Original} & Swahili & 0.000000 &  &  &  & \\
			& ShiNgazidja & 41.112070 & 0.000000 &  &  &  \\
			& ShiMaore & 46.197453 & 17.570644 & 0.000000 &  &  \\
			& ShiNdzuani & 47.975246 & 12.963908 & 4.271070 & 0.000000 &  \\
			& ShiMwali & 49.700855 & 8.815629 & 11.465710 & 9.981685 & 0.000000 \\
			\midrule
			\multirow{5}{*}{Swadesh 200} & Swahili & 0.000000 &  &  &  &  \\
			& ShiNgazidja & 42.132576 & 0.000000 &  &  &  \\
			& ShiMaore & 46.932243 & 17.292781 & 0.000000 &  &  \\
			& ShiNdzuani & 48.094538 & 13.027077 & 4.250638 & 0.000000 &  \\
			& ShiMwali & 48.831501 & 9.427224 & 8.787436 & 7.045041 & 0.000000 \\
			\midrule
			\multirow{5}{*}{Swadesh-Yakhontov} & Swahili & 0.000000 &  &  &  &  \\
			& ShiNgazidja & 36.957672 & 0.000000 &  &  &  \\
			& ShiMaore & 37.303922 & 17.850140 & 0.000000 &  & \\
			& ShiNdzuani & 40.483871 & 10.476190 & 4.462366 & 0.000000 &  \\
			& ShiMwali & 43.222222 & 2.738095 & 11.444444 & 10.388889 & 0.000000 \\
			\bottomrule
		\end{tabular}
	\end{center}
	\label{tab:distance}
\end{table*}

\section{Main Contributions} \label{datasets}
\subsection{Datasets}
Obtaining sufficient and high-quality data to train NLP solutions is often a tedious task given the costs involved. The most intuitive approach is to hire individuals to contribute to the labeling of large amounts of data and ensure their reliability, as was the case for ChatGPT \cite{Ray2023}. Such an approach requires significant funding, considerable human resources, and other resources.

For the case of the previous example, it was mainly about a proprietary initiative where the data belongs to a specific entity. As for the CommonVoice initiative \cite{ardila2020common}, the approach is slightly different. Although it requires a large mobilization of human resources, the financial costs are lower since participants contribute for free to build the largest speech datasets possible. However, it still remains quite costly in terms of time, as it requires volunteers to come and record their voices by reading snippets of text in a chosen language.

Although the approaches we have just mentioned are among the most reliable for obtaining high-quality data, it is not easy for everyone to implement them. This has largely necessitated the use of other methods such as synthetic data, data augmentation, or pseudo-labeling \cite{lee2023datadriven, fazel2021synthasr, REBAI2017316, naira-etal-2024-datasets}. In our case, where no funding was allocated for this work, we had to resort to pseudo-labeling solutions to construct two datasets, for MT and for ASR. The objective of this work is therefore to demonstrate that with limited resources, one can significantly contribute to improving the state-of-the-art in NLP for an extremely low-resource language.

\subsubsection{Machine Translation} \label{mt_data}
The datasets found in \cite{naira-etal-2024-datasets} are the starting points here. 17,000 documents (complete sequences and dictionaries) in Comorian translated into French and English are still very few if one wants to achieve a highly efficient translation system. Besides the small size of the data, other factors could influence the model's performance. In our case, a large portion of the sequences comes from the Bible and documents extracted from the Jehovah's Witnesses portal\footnote{\url{https://www.jw.org/en/}}. In fact, one thing to know here is that having data from a specific domain is not problematic in itself because it allows for a better understanding of that domain \cite{martins2022efficient}. The problem arises when one wishes to generalize the model across multiple domains. In that case, it is important to consider creating the most heterogeneous dataset possible. This made it necessary to resort to other solutions to diversify the data. This diversification could also lead to significant data volumes.

\begin{figure}[ht]
	\centering
	\includegraphics[width=0.6\linewidth]{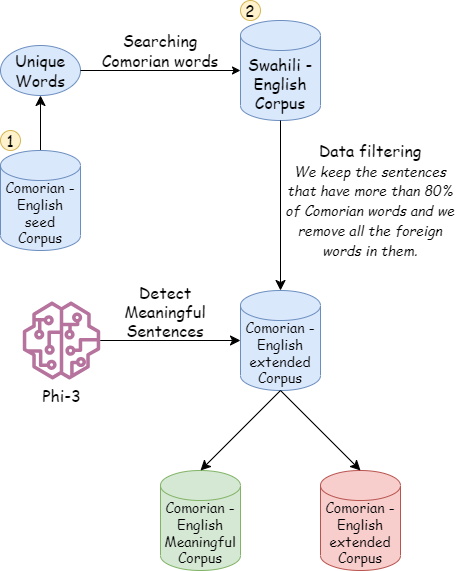}
	\caption{NMT Data Construction.}
	\label{fig:nmt_data}
\end{figure}

Previous works \cite{Hujon2023, aji-etal-2020-neural} have demonstrated the importance of transfer learning in improving MT systems for low-resource languages. The proposed solutions involve using a well-resource and closely related intermediary language to build a model that understands the low-resource one. In our case, we propose to follow the pipeline outlined in Figure \ref{fig:nmt_data} to create an automatic translation dataset from Comorian to English using Swahili as the intermediary language.

The construction of the dataset involves several steps. The first step is to gather various corpora in Swahili. These corpora do not necessarily need to be translated into other languages; raw texts are sufficient. In the Comorian dataset, we then extract the unique words from the entire corpus. After that, we return to the Swahili corpus and count the Comorian words in each sentence. Finally, we retain sentences with at least 80\% Comorian words. To translate these Swahili sentences into English, we use the DeepTranslator package\footnote{\url{https://pypi.org/project/deep-translator/}}. The final dataset consists of 30,000 translated sentences along with the original 17,000 Comorian data points.

\subsubsection{Speech Recognition}
For ASR, we employ an approach quite similar to the one used for MT. We download the Swahili segments of the CommonVoice dataset\footnote{\url{https://commonvoice.mozilla.org/fr/datasets}}, filtering it based on the Comorian words obtained earlier. The difference lies in our method: instead of seeking exact matches, we initially compute the Levenshtein distances between each unique Swahili word and the Comorian words. Subsequently, we identify texts in the CommonVoice dataset where at least 80\% of the words contain at least one Comorian word with a Levenshtein distance of 80\% or more. Ultimately, we acquire 4 hours of labeled data.

\subsection{Models}

All our experiments were conducted on a T4 GPU in Google Colab\footnote{\url{https://colab.research.google.com/}}. While the architectures used necessitate GPU acceleration for swift and efficient training, the constraints of available resources (only 12 GB of memory) limited us to utilizing smaller checkpoints of the models, such as mt5-small or whisper-small.

\subsubsection{Multilingual T5}
mT5 \cite{xue-etal-2021-mt5} is a multilingual variant of T5 \cite{raffel2023exploringlimitstransferlearning}, a language model developed by Google Research. Designed to preserve the core characteristics of T5, mT5 adopts a versatile text-to-text format and is built upon insights gained from a large-scale empirical study. This model is trained on mC4\footnote{\url{https://huggingface.co/datasets/allenai/c4}}, a multilingual dataset comprising texts in 101 languages sourced from CommonCrawl. mT5 excels across various NLP tasks, demonstrating state-of-the-art performance on multiple benchmark datasets. A significant advancement of mT5 is its capability to mitigate inadvertent translation errors in zero-shot predictions through the integration of unlabeled data during fine-tuning.

The ROUGE scores (see Table \ref{tab:mt5}) of the machine translation model are promising for an initial evaluation. With ROUGE-1 at 0.6826, ROUGE-2 at 0.42, ROUGE-L at 0.6532, and ROUGE-Lsum at 0.6533, the model demonstrates an ability to capture similarity and recall between reference and translation predictions. These results suggest a good fit in reproducing relevant sentence structures and linguistic features compared to the reference data, indicating potential for improved performance with additional training iterations and model adjustments.

\begin{table*}[h]
	\footnotesize
	\caption{NMT Results.}
	\begin{center}
		\begin{tabular}{@{}llllll@{}}
			\toprule
			
			Checkpoint & Rouge1 & Rouge2 & Rougel & Rougelsum & Gen Len \\
			\midrule
			\href{https://huggingface.co/nairaxo/english-shikomori-nmt}{nairaxo/english-shikomori-nmt} & 0.6826 & 0.42 & 0.6532 & 0.6533 & 11.492 \\
			\bottomrule
		\end{tabular}
	\end{center}
	\label{tab:mt5}
\end{table*}

\subsubsection{Whisper}
Whisper \cite{radford2022robustspeechrecognitionlargescale} employs an encoder-decoder Transformer architecture for speech recognition, with all audios resampled to 16,000 Hz and converted into 80-channel Mel spectrograms. The encoder processes this input through convolution layers and Transformer blocks, while the decoder uses the learned token representations and the positional embeddings. The model utilizes a byte-level BPE tokenizer for both English and multilingual models. Additionally, Whisper adopts a multitask format, allowing a single model to handle various speech processing tasks by specifying these tasks through sequences of input tokens to the decoder.

The results (see Table \ref{tab:whisper}) of the speech recognition model indicate notable performance with a loss of 0.86, a Word Error Rate (WER) of 39.50\%, and a Character Error Rate (CER) of 13.76\%. The WER and CER, although non-negligible, suggest that the model manages to recognize and transcribe text with reasonable accuracy, while highlighting opportunities for improvement in reducing transcription errors.

\begin{table*}[h]
	\footnotesize
	\caption{ASR Results.}
	\begin{center}
		\begin{tabular}{@{}lll@{}}
			\toprule
			
			Checkpoint & WER & CER \\
			\midrule
			\href{https://huggingface.co/nairaxo/asr-shikomori-swahili}{nairaxo/asr-shikomori-swahili} & 39.50 & 13.76 \\
			\bottomrule
		\end{tabular}
	\end{center}
	\label{tab:whisper}
\end{table*}


\section{Conclusion} \label{conclusion}
In light of these experiments, it is clear that low-resource African languages such as Comorian benefit minimally from current advancements in NLP. While languages like Swahili have sufficient resources to develop high-performing systems, the situation is markedly different for many other languages across the Africa. In this context, we explored transfer learning as a promising solution to address this critical lack of data. By using Swahili as a source language and applying our approach to Comorian, we initiated the development of NLP technologies for these emerging languages. Our initial results in speech recognition and machine translation demonstrate the feasibility of this approach, paving the way for broader adoption of these beneficial technologies in diverse contexts.

\bibliographystyle{abbrvnat}
\bibliography{main}

\end{document}